\documentclass{article}
\usepackage[preprint]{neurips_2025}
\usepackage[utf8]{inputenc} 
\usepackage[T1]{fontenc}    
\usepackage{hyperref}       
\usepackage{url}            
\usepackage{booktabs}       
\usepackage{amsfonts}       
\usepackage{nicefrac}       
\usepackage{microtype}      
\usepackage{xcolor}         
\usepackage{amsmath}
\usepackage{amssymb, amsmath}

\usepackage{subfig}
\usepackage{graphicx}
\usepackage{siunitx}
\usepackage{amsthm}

\newtheorem{theorem}{Theorem}

\title{Modeling Adoptive Cell Therapy in Bladder Cancer from Sparse Biological Data using PINNs }
\author{
 Kayode Olumoyin \\
 H. Lee Moffitt Cancer Center and Research Institute\\
  Integrated Mathematical Oncology Department\\
  Tampa, FL \\
  \texttt{kayode.olumoyin@moffitt.org} \\
   \And
 Katarzyna Rejniak \\
 H. Lee Moffitt Cancer Center and Research Institute\\
  Integrated Mathematical Oncology Department\\
  Tampa, FL \\
  \texttt{kasia.rejniak@moffitt.org} \\
}
\begin{document}
\maketitle
\begin{abstract}
Physics-informed neural networks (PINNs) are neural networks that embed the laws of dynamical systems modeled by differential equations into their loss function as constraints. In this work, we present a PINN framework applied to oncology. Here, we seek to learn time-varying interactions due to a combination therapy in a tumor microenvironment. In oncology, experimental data are often sparse and composed of a few time points of tumor volume.  By embedding inductive biases derived from prior information about a dynamical system, we extend the physics-informed neural networks (PINN) and incorporate observed biological constraints as regularization agents. The modified PINN algorithm is able to steer itself to a reasonable solution and can generalize well with only a few training examples. We demonstrate the merit of our approach by learning the dynamics of treatment applied intermittently in an ordinary differential equation (ODE) model of a combination therapy. The algorithm yields a solution to the ODE and time-varying forms of some of the ODE model parameters.  We demonstrate a strong convergence using metrics such as the mean squared error (MSE), mean absolute error (MAE), and mean absolute percentage error (MAPE).
\end{abstract}
%
\section{Introduction}
%
In a biological model of interacting organisms in their environment, where internal signals or external interventions may alter behavior over time, model parameters such as growth and migration rates may be time-varying~\citep{benaim2024, muller2019}. For instance, in chemotherapy and immunotherapy models, where we want to maximize tumor kill while minimizing side effects~\citep{galluzzi2015, emens2016}, intervention strategies such as drug dosing schedules may cause model parameters to vary over time. This suggests that modeling the growth rate parameter as time-varying may provide a more accurate representation of interaction dynamics. This motivates the need to develop algorithms that accurately model real-world dynamics where the underlying conditions are also evolving. So if data suggest that fixed parameters cannot capture the observed dynamics, allowing parameters to vary with time may provide flexibility for better model fitting. Another motivation for time-varying parameters for biological models is that, in some cases, we lack full mechanistic knowledge of the model. Here, time-varying parameters behave similar to equation learning algorithms such as symbolic regression and neural ordinary differential equations (NODEs)~\citep{udrescu2020sciAdv,chen2018neurips} which can act as surrogates for unobserved or unmodeled effects~\citep{glyndavies2025, sawada2022}.
%

In differential equations model of a biological system, data are often sparse because: \((1)\) they are collected at irregular time points; \((2)\) the collected data may contain noise due to measurement errors; and \((3)\) the interacting variables may only be partially observable. Data augmentation methods are often needed to extend the sparse available data, but the choice of the right data augmentation method depends on factors such as mitigating overfitting~\citep{mumuni2022array} and improving model performance~\citep{svabensky2025acm}. Spline-based interpolation is a powerful data augmentation approach to modeling smooth, continuous time-varying parameters between sparse data points, especially in biological systems where the assumption of gradual change is often biologically reasonable~\citep{zhan2014}.
%

Advancements in machine learning and data science have made it possible to learn complex dynamics directly from data. Data-driven approaches, including neural networks and nonlinear regression, have been succesfully applied to dynamical systems~\citep{bongard2007dyn, brunton2016sindy, raissi2018multinet}. For instance, in~\citep{brunton2016sindy}, they introduced a method called sparse identification of nonlinear dynamical systems (SINDy) to extract governing equations from data. They demonstrate that SINDy can learn nonlinear systems from data, but requires some prior knowledge of the system’s dynamics in order to construct a suitable library of candidate functions. In~\citep{brummer2023fimmu}, the authors demonstrate the potential of an extension to SINDy for sparse data. However, like in many other variants of the SINDy algorithm on sparse data, performance degrades in limited-data scenario~\citep{fasel2022rspa}. In our case, where we do not have data for all of the observed variables, we want a method that use available data and known dynamics as constraints to learn admissible solution to a system of differential equations. One of the most successful approaches in this regard in recent years is the physics-informed neural networks (PINNs), introduced in~\citep{raissi2019pinn}. One of the ways we can use PINNs is called the inverse problem, where we assume the form of the differential equation to be known and the goal is to learn the differential equation parameters from data. 
%
%

Many recent advancements in deep learning have been influenced by PINNs, leading to algorithms that have been successfully applied to learn model parameters and update governing equations from data~\citep{lagergren2020binn, podina2024pinn}. However, these methods do not perform well in a limited data scenario, especially when interacting subpopulations data are not known. The original PINN algorithm~\citep{raissi2019pinn} allows for various modifications, and researchers have explored different approaches such as adaptive learning rates~\cite{fan2024pinn}, adaptive loss weighting~\citep{mcclenny2022pinn, gao2025pinn}, and time-varying differential equation parameters~\citep{kharazmi2021pinn, olumoyin2021pinn}.
%

What is often lacking, however, is a method for learning time-varying parameters in systems of differential equations when the experimental training data represent the sum of multiple subpopulations at a limited number of time points, while the individual subpopulation data remain unknown. In this work, we propose a modification to the PINN algorithm that captures unmodeled effects directly from the observed data. This approach extends the differential equation model and remains robust even in limited-data scenarios.

We claim the following contributions:
\begin{enumerate}
	\item We propose a novel neural network approach that modifies PINN to learn unmodeled effects from data in a limited data scenario
	\item We develop a framework that allows the governing equations, experimental data, and biological constraints to act as regularization agents that together constrain the space of admissible solutions.
	\item Our approach is able to encode prior information about the dynamics which enables us to discard non-realistic solutions.
	\item We are able to update governing equations from limited data, which provides a more accurate representation of observed dynamics.
\end{enumerate}
%
%
%
%
%
%
%
%
\section{Methods}
\subsection{Problem Statement}
Consider a system of ODEs in the following form, with \(m\) variables and \(n\) parameters, where at most \((p \leq n)\) of the parameters are time-varying.
\begin{equation}\label{orig_eqn}
\frac{du(t)}{dt} = f(t, u(t);(\lambda_1, \lambda_2, \ldots, \lambda_n)), 
\end{equation}
where \(u: t \rightarrow {\rm I\!R}^m\), \(t\in[t_0,t_F]\), and \(\Lambda(t) = (\lambda_1(t), \lambda_2(t), \ldots, \lambda_p(t))\) denote \(p\) unknown time-varying parameters. The function \(f\) is nonlinear. Then, given \(M^{\text{data}}\) experimental data points \(\{(t_i, u(t_i)) \}_{i=0}^{M^{\text{data}}-1}\) which approximately satisfies Eq.~\eqref{orig_eqn}, we can model \(u\) and \(\Lambda\) using neural network surrogates denoted \(u_{NN}\) and \(\Lambda_{NN}\) respectively. So that Eq.~\eqref{orig_eqn} can be written as follows:
\begin{equation}\label{orig_eqn2}
\frac{d u_{NN}(t)}{dt} = f(t, u_{NN}(t);({\lambda_{NN}}_1, {\lambda_{NN}}_2, \ldots, {\lambda_{NN}}_n)), 
\end{equation}
%
%
and \(\Lambda_{NN}(t) = ({\lambda_{NN}}_1(t), {\lambda_{NN}}_2(t), \ldots, {\lambda_{NN}}_p(t))\). Each of \(u_{NN}\) and \(\Lambda_{NN}\) can be learned as the outputs of a Feedforward Neural Network (FNN).
\subsection{Feedforward Neural Network (FNN)} \label{subSEC_FNN}
An FNN is an artificial neural network where the forward pass is in one direction, and it is usually fully connected. It can be represented as a composition of functions with \(L\) layers, where the input layer is an affine transformation of a vector \(t\) followed by an activation function. The output of the input layer becomes input to the next layer, continuing until the \(L\)th layer as shown in Eq.~\eqref{fnneq}.  
 \begin{equation}\label{fnneq}
 NN(t;\theta) = \sigma_{L}(W_{L}\sigma_{L-1}(W_{L-1}\ldots \sigma_{2}(W_{2} \sigma_{1}(W_{1}t + b_1)+b_2)\ldots+b_{L-1})+b_L),
 \end{equation}
 \noindent where \(\theta : = (W_1, \ldots, W_L,b_1, \ldots, b_L)\) denotes the set of neural network weight matrices \(W_k\) and bias vectors \(b_k\) for \((k = 1, \ldots, L)\), and \(\sigma_k (\cdot)\) denotes the activation function. It is widely known that a feedforward neural network with at least one hidden layer containing a finite number of neurons can approximate any continuous function on compact subsets of \({\rm I\!R}^m\)~\citep{hornik1991}. 
\subsection{Modified PINN}\label{subSEC_mPINN}
In a limited data scenario of \(M^{\text{data}}\) experimental data points \(\{(t_i, u(t_i)) \}_{i=0}^{M^{\text{data}}-1}\), we can generate \(M^{\text{interp}}\) additional points which increases the experimental data points within its domain. We provide a formal description of this data extension approach using spline-based interpolation on the PINN solution and time-varying parameters in Eq.~\eqref{orig_eqn2}.
\begin{theorem}[Interpolation preserves convergence of ODE solution in Eq.~\eqref{orig_eqn2}]
\label{thm:interp-ode}
Suppose that:
\begin{enumerate}
  \item[(i)] \(u,u_{NN}\in C^{k+1}([t_0,t_F];\mathbb{R}^m)\) are functions satisfying Eq.~\eqref{orig_eqn} and Eq.~\eqref{orig_eqn2}, respectively.
  \item[(ii)] The PINN surrogate parameters satisfy \(\Lambda_{NN} \to \Lambda\) uniformly on \([t_0,t_F]\), and the PINN surrogate solution satisfies
  \[
  u_{NN}\to u \quad \text{in } C^{k}([t_0,t_F];\mathbb{R}^m).
  \]
  \item[(iii)] \(I_k\) is the spline interpolation operator of degree \(k\) on a quasi-uniform mesh on \([t_0,t_F]\) with spacing \(h>0\).
  \item[(iv)] The constant $C_{\text{interp}}$ denotes the spline interpolation error constant such that
  \[
  \|I_k v - v\|_{L^\infty(t_0,t_F)} \le C_{\text{interp}} h^{k+1} \|v^{(k+1)}\|_{L^\infty(t_0,t_F)}
  \]
  for any $v \in C^{k+1}([t_0,t_F];\mathbb{R}^m)$.
\end{enumerate}
Then
\[
\|I_k u_{NN} - u\|_{L^\infty(t_0,t_F)} 
\le \|I_k u_{NN} - I_k u\|_{L^\infty(t_0,t_F)} + \|I_k u - u\|_{L^\infty(t_0,t_F)}.
\]
Moreover, as $h \to 0$:
\[
\|I_k u_{NN} - u\|_{L^\infty(t_0,t_F)} \to 0.
\]
\end{theorem}
%
%
%
%
%
We formulate two PINNs using the FNN setup described in Section~\ref{subSEC_FNN}. Both PINNs take as input the time domain \(t \in [t_0, t_F]\), corresponding to the interpolation function \(\hat{u}(t)\) that interpolates the data \(\{(t_i, u(t_i))\}\). By Theorem~\eqref{thm:interp-ode}, the output of the first PINN, denoted \(\hat{u}_{NN}(t)\), which satisfies Eq.~\eqref{orig_eqn3}, is an admissible approximation to the ODE solution in Eq.~\eqref{orig_eqn2}. In the theoretical framework, the spline interpolant \(I_ku_{NN}\) serves as an analytical surrogate for 
\(\hat{u}_{NN}\). In practice, however, \(\hat{u}_{NN}\) itself plays the role of this interpolant, offering a smooth, differentiable representation over \([t_0, t_F]\) that satisfies the learned dynamics:
\begin{equation}\label{orig_eqn3}
\frac{d \hat{u}_{NN}(t)}{dt} 
= f\big(t, \hat{u}_{NN}(t); (\Lambda_{NN}(t), {\lambda_{NN}}_{p+1}, \ldots, {\lambda_{NN}}_n)\big).
\end{equation}
The second PINN models the time-varying parameters \(\Lambda(t)\), so that PINN learns a mapping \(\Lambda_{NN}(t) \approx \Lambda(t)\) over the same interval \([t_0, t_F]\), capturing temporal variations that cannot be represented by constant parameters. Together, the coupled PINNs form a hierarchical system in which \(\Lambda_{NN}(t)\) helps to understand the evolution of 
\(\hat{u}_{NN}(t)\), and Theorem~\eqref{thm:interp-ode} ensures that, under convergence of both networks and sufficiently fine interpolation, 
the composite approximation converges uniformly to the true ODE solution \(u(t)\).
%
%

The loss function in Eq.~\eqref{loss1} is used to train both PINNs. The output of the first PINN \(\hat{u}_{NN}\) enables us to learn subpopulation dynamics from data, while the output of the second PINN, \(\Lambda_{NN}\), allows us to infer interaction dynamics from the data. In both PINNs, we implement the activation function \(\sigma_k (\cdot)\) (Eq.~\eqref{fnneq}) using the \verb+tanh+ activation function in the hidden layers and the \verb+softplus+ activation function in the output layers. This setup constrains the PINN to produce only non-negative solutions. We initialize the weight matrices and bias vectors  \((W_1, \ldots, W_L,b_1, \ldots, b_L)\) using \verb+Glorot initialization+. Optimization was performed using the \verb+Adams optimizer+ with \(20,000\) \verb+Epochs+ to ensure loss convergence. For the PINN that learns \(\hat{u}_{NN}\),  we use \(3\) hidden layers with \(100\) neurons per each hidden layer. For the PINN that learns \(\Lambda_{NN}\), we use \(2\) hidden layers with \(200\) neurons per each hidden layer. All implementation was carried out using the \verb+TensorFlow+ library.
\begin{equation}\label{loss1}
	\begin{split}
		\mathcal{L} = w_{d} \mathcal{L}_{d} + w_{IC} \mathcal{L}_{IC} + w_{r} \mathcal{L}_{r} + w_{bc} \mathcal{L}_{bc}
        \end{split}
\end{equation}
The loss \(\mathcal{L}\) is defined as a weighted sum of the norms of the observed data loss  (\(\mathcal{L}_{d}\)), the initial conditions loss (\(\mathcal{L}_{IC}\)), the ODE residual loss  (\(\mathcal{L}_{r}\)), and biology constraint loss (\(\mathcal{L}_{bc}\)). The weights \( w_{d}\), \( w_{IC}\), \( w_{r}\), \( w_{bc}\) are manually adjusted during the implementation of the algorithm in order to address imbalances in the magnitude of the different loss components in Eq.~\eqref{loss1}. The complete loss function \(\mathcal{L}\) is defined in Eq.~\eqref{loss2} and \(\|\cdot\|\) is an arbitrary norm.
%
%
\begin{equation}\label{loss2}
	\begin{split}
		\mathcal{L}_{d} &= \frac{1}{M^{\text{data}}}\sum_{i = 0}^{M^{\text{data}}-1} \| \hat{u}_{NN}(t_i) - \hat{u}(t_i)  \| \\
		\mathcal{L}_{IC} &= \|  \hat{u}_{NN}(t_0) - \hat{u}(t_0)\| \\
		\mathcal{L}_{r} &= \frac{1}{M^{\text{data}}+M^{\text{interp}}}\sum_{j = 0}^{M^{\text{data}}-1+M^{\text{interp}}} \| \frac{d} {d t_j} \hat{u}_{NN}(t_j) - f(t_j, \hat{u}_{NN}(t_j);(\Lambda_{NN}(t_j),{\lambda_{NN}}_{p+1}, \ldots, {\lambda_{NN}}_{n})) \|\\
		\mathcal{L}_{bc} &= \frac{1}{M^{\text{bc}}}\sum_{k =1,\ldots, M^{\text{bc}}} \| \hat{u}_{NN}(t_k) - \hat{u}(t_k)  \| \\
        \end{split}
\end{equation}
In Eq.~\eqref{loss2}, we compute \(\frac{d} {d t_j} \hat{u}_{NN}\) using the automatic differentiation package as described in~\citep{baydin2018} and evaluate at the time points \(t_j\). The data loss \(\mathcal{L}_{d}\) penalizes \(\hat{u}_{NN}\) until it matches the experimental data \(u(t_i)\).  The residual loss \(\mathcal{L}_{r}\) penalizes \(\Lambda_{NN}\) until it matches the differential equation in Eq.~\eqref{orig_eqn3} at the experimental and interpolation point \(M^{\text{data}}+M^{\text{interp}}\). It is important to note that forcing the neural network to satisfy the initial conditions closely is beneficial; this enforcement is achieved through \(\mathcal{L}_{IC}\). The loss term \(\mathcal{L}_{bc}\) is used to incorporate additional biological constraints on the subpopulations at a small number of time points, denoted \(M^{\text{bc}}\).
\section{Results}
\subsection{ODE Model for a Combination Therapy}\label{chemimmu}
%
We describe an ODE model of a growing tumor exposed to a combination of two anticancer treatments: gemcitabine (GEM) chemotherapy and T cell immunotherapy, based on a study published in \citep{bazargan2023til}. In this experimental study, the \SI{1e5} MB49-OVA bladder cancer cells were instilled orthotopically into the mouse bladder. This was followed on day \(10\) by intravesical injection of \(500 \mu g\) of GEM to locally deplete the tumor microenvironment from the immunosuppressive cells, such as the myeloid-derived suppressive cells (MDSCs).  Next, at day \(14\), the \SI{5e6} OT-1 T cells were administered intravesically into the mouse bladder. The OT-1 T cells are engineered to recognize the MB49-OVA tumor. This immunotherapy treatment is a pre-clinical analogue of the adoptive cell therapy with tumor-infiltrating lymphocytes (ACT-TIL), that is a personalized immunotherapy using patients’ own T cells expanded ex vivo and reinfused to the patient \citep{rosenberg2015til}. In the pre-clinical study, tumor overall volume was monitored using ultrasound imaging and recorded on days \(6, 9, 13, 16, 20\) and \(23\).
We propose the following systems of ODEs to describe the interactions between cancer cells \( (C)\), T cells \( (T)\), MDSCs cells \( (M)\), and GEM \( (G)\). These are presented in Eqs~\eqref{gemot1eqn}, with the summary of the model parameters presented in Table~\eqref{paramtable}. 
\begin{equation}\label{gemot1eqn}
	\begin{split}
		\frac{dC}{dt} &= p_C C - k_{TC} CT - k_{GC} CG\\
		\frac{dT}{dt} &= U_T + n_T T - s_{CT}TC - s_{MT} TM - k_{GT}TG\\
		\frac{dM}{dt} &= r_M C - k_{GM}MG - d_{M} M\\
		\frac{dG}{dt} &= U_G  - d_{G} G
        \end{split}
\end{equation}
Both chemo- and immune-therapy can affect the interactions between cancer cells, T cells, and MDSCs. To investigate how the trajectories of these altered dynamics can be inferred from the experimental data, we assume that the following three model parameters can vary over time: the proliferation rate of cancer cells \((p_C)\), the death rate of the MDSCs \((d_M)\), and the recruitment rate of the MDSCs \((r_M)\). 
\\
%
%
\begin{table}[h]
  \caption{Table of parameters in Eq. ~\eqref{gemot1eqn} }
  \label{paramtable}
  \centering
  \resizebox{0.55\textwidth}{!}{%
  \begin{tabular}{ll}
    \toprule
    Parameters     & units      \\
    \midrule
    $k_{TC}$ -- Rate of Cancer cells killed by T cells   & $mm^{-3} day^{-1}$     \\
    $k_{GC}$ -- Rate of Cancer cells killed by GEM    & $mg^{-3} day^{-1}$     \\
    $k_{GM}$ -- Rate of MDSCs killed by GEM    & $mg^{-3} day^{-1}$     \\
    $k_{GT}$ -- Rate of T cells killed by GEM    & $mg^{-3} day^{-1}$     \\
    $d_{G}$ -- GEM decay/clearance rate    & $day^{-1}$     \\
    $d_{M}$ -- MDSCs cell death rate    & $day^{-1}$     \\
    $p_{C}$ -- Cancer cell proliferation rate    & $day^{-1}$     \\
    $n_{T}$ -- T cell net growth rate    & $day^{-1}$     \\
    $U_{G}$ -- GEM injection schedule    & $mg $     \\
    $U_{T}$ -- T cell injection schedule    & $mm^{3} $     \\
    $r_{M}$ -- MDSCs recruitment rate    & $ day^{-1}$     \\
    $s_{MT}$ -- MDSCs suppression rate    & $mm^{-3} day^{-1}$     \\
    $s_{CT}$ -- Cancer cell suppression rate    & $mm^{-3} day^{-1}$     \\
    \bottomrule
  \end{tabular}}
\end{table}
\subsection{Application of the Modified PINN to the Combination Therapy }\label{chemimmu2}
\begin{figure}[h]
  \includegraphics[width = 5.5in]{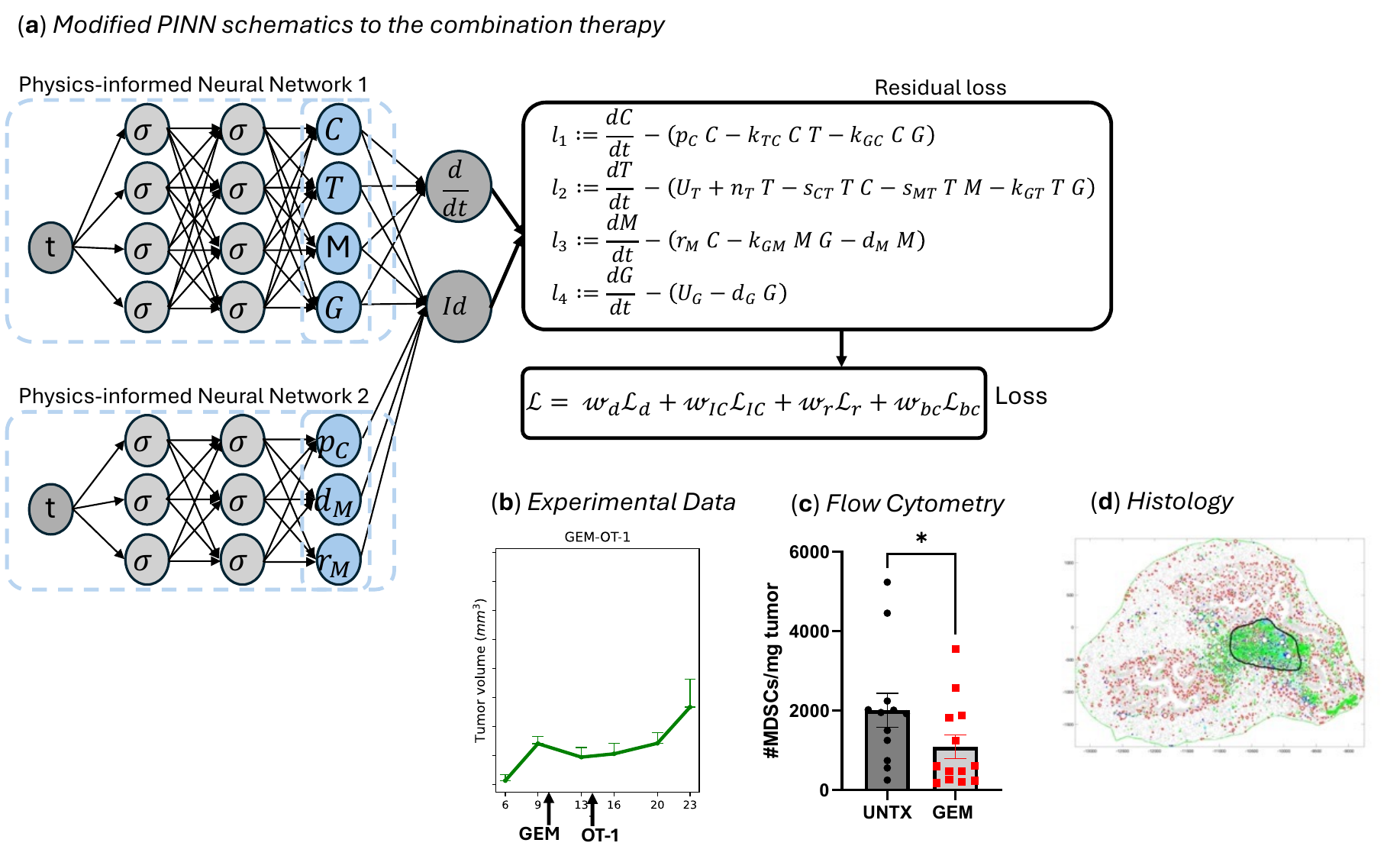}
  \caption{Modified PINN applied to the system of ODEs in Eq.~\eqref{gemot1eqn}}
  \label{expdata}
\end{figure}

We will demonstrate our approach by applying the modified PINN (section~\eqref{subSEC_mPINN}) to an ODEs model of a growing tumor exposed to a combination of two anticancer treatments: gemcitabine (GEM) chemotherapy and T cell immunotherapy described in section~\eqref{chemimmu}. In Figure~\eqref{expdata}({\bf a}), we present the schematic of the implementation of the PINN algorithm to the system of ODEs given in Eq.~\eqref{gemot1eqn}. There are two PINNs in Figure~\eqref{expdata}({\bf a}), the first PINN learns the dynamics of the subpopulations \( C\), \( T\), \( M\), and \( G\), which are potential solution to Eq.~\eqref{gemot1eqn}. The second PINN learns the three time-varying parameters \(p_C\), \(d_M\), and \(r_M\) in an unsupervised way to infer interaction dynamics resulting from the external interventions of GEM (chemotherapy) and OT-1 T cells (immunotherapy). We denote the experimental data as \(u_{GEM-OT1}\), which are tumor volume measurements from ultrasound images recorded on days \(6, 9, 13, 16, 20\), and \(23\), see Figure~\eqref{expdata}({\bf b}). The arrow markers in Figure~\eqref{expdata}({\bf b}) indicate the intravesical injection of GEM and OT-1 T cells at the time points \(10\) and \(14\), respectively. Now, we assume that \(u_{GEM-OT1}\) is a sum of the subpopulations \( C\), \( T\), and \( M\), so we implement the data loss \(\mathcal{L}_{d}\) in Eq.~\eqref{loss2} as \( \frac{1}{M^{\text{data}}}\sum_{i = 0} ^ {M^{\text{data}}-1} ((C(t_i) + T(t_i) + M(t_i)) - u_{GEM-OT1}(t_i)  )^{2}  \). 
%

In order to initialize \( C\), \( T\), \( M\), at \(t_0 = 6\), we define \(\mathcal{L}_{_{IC}}\) in Eq.~\eqref{loss2} as follows:  
\begin{equation*}\label{loss221}
	\begin{split}
	\mathcal{L}_{IC} &= \left( C(t_0) - p_{0_{C}} \times u_{GEM-OT1}(t_0)  \right)^{2}  +  \left( T(t_0) - p_{0_{T}} \times u_{GEM-OT1}(t_0)  \right)^{2} \\
	&+  \left( M(t_0) - p_{0_{M}} \times u_{GEM-OT1}(t_0)  \right)^{2},
	\end{split}
\end{equation*}
where we obtain the proportions of \(C\) and \(M\) at \(t_0 = 6\) from flow cytometry data (Figure~\eqref{expdata}({\bf c})), so that \(p_{0_{C}} = 0.99887\), \(p_{0_{T}} = 0\), and \(p_{0_{M}} = 1 - 0.99887\). 

The residual loss \(\mathcal{L}_{r}\) in Eq.~\eqref{loss2} for the system of ODEs in Eq.~\eqref{gemot1eqn} is as follows:
\begin{equation*}\label{loss222}
	\begin{split}
 	\mathcal{L}_{r} &= \sum_{i = 1} ^ 4 \left(\frac{1}{M^{\text{data}}+M^{\text{interp}}}\sum_{j = 0}^{M^{\text{data}}-1+M^{\text{interp}}} l_{i}(t_j) ^{2}\right),
 	\end{split}
\end{equation*}
and 
\begin{equation*}\label{loss223}
	\begin{split}
 	l_1 &= \frac{dC}{dt} - \left( p_C C - k_{TC} C T - k_{GC} C G \right)\\
	l_2 &= \frac{dT}{dt} - \left( U_T + n_T T - s_{CT} T C - s_{MT} T M - k_{GT} T G \right)\\
	l_3 &= \frac{dM}{dt} - \left( r_M C - k_{GM} M G - d_M M \right)\\
	l_4 &= \frac{dG}{dt} - \left( U_ G - d_{G} G \right).
 	\end{split}
\end{equation*}
In Figure~\eqref{expdata}({\bf d}), we present a digital histology of bladder tissue from a mouse treated with GEM and OT-1 T cells. Tumors \((C)\) are outlined in black, MDSCs \((M)\) are represented as green dots, and T cells \((T)\) are blue dots. Using two histology time points data collected on days \(t_{k1} = 17\) and \(t_{k2} = 23\), we implement the biology constraints loss \(\mathcal{L}_{bc}\) in Eq.~\eqref{loss2} as follows:
\begin{equation*}\label{loss224}
	\begin{split}
	\mathcal{L}_{bc} &= \left(  C(t_{k1}) - p_{{k1}_{C}} \times u_{GEM-OT1}(t_{k1}) \right) ^{2}  + \left(  T(t_{k1}) - p_{{k1}_{T}} \times u_{GEM-OT1}(t_{k1}) \right) ^{2}  \\
	&+ \left(  M(t_{k1}) - p_{{k1}_{M}} \times u_{GEM-OT1}(t_{k1}) \right) ^{2} \\
	&+ \left(  C(t_{k2}) - p_{{k2}_{C}} \times u_{GEM-OT1}(t_{k2}) \right) ^{2}  + \left(  T(t_{k2}) - p_{{k2}_{T}} \times u_{GEM-OT1}(t_{k2}) \right) ^{2}  \\
	&+ \left(  M(t_{k2}) - p_{{k2}_{M}} \times u_{GEM-OT1}(t_{k2}) \right) ^{2} .
	\end{split}
\end{equation*}
Where \(p_{{k1}_{C}} = 0.95755\), \(p_{{k1}_{T}} = 0.01818\), \(p_{{k1}_{M}} = 0.02427\) are the proportions of \( C\), \( T\), and \( M\) respectively at day \(t_{k1} = 17\) (histology time point).  Similarly for \( C\), \( T\), and \( M\) at day \(t_{k2} = 23\) (histology time point), we have that \(p_{{k2}_{C}} = 0.95665\), \(p_{{k2}_{T}} = 0.00078\), and \(p_{{k2}_{M}} = 0.04256\) respectively.

Because the experimental data \(u_{GEM-OT1}\) are quite sparse, and are assumed to be the sum of the subpopulations \( C\), \( T\), and \( M\), which are unknown at the experimental time points, deep learning methods will fail to generalize to new data point if trained only on these limited data. We address this challenge by increasing the number of data points within the points where the experimental data points were collected using spline-based interpolation. It is not formally known how many \(M^{\text{interp}}\) data points are needed to extend the size of \(u_{GEM-OT1}\) from \(M^{\text{data}}\) to \(M^{\text{data}} + M^{\text{interp}}\). To investigate this, we will consider two cases: case \((1)\) we take \(M^{\text{data}} + M^{\text{interp}} = 86\) data points, where we generate \(M^{\text{interp}} = 80\) uniformly spaced points between the \(M^{\text{data}} = 6\) experimental data points. And case \((2)\) here, \(M^{\text{data}} + M^{\text{interp}} = 35\) data points of which \(M^{\text{interp}} = 29\) uniformly spaced interpolation points between the experimental data. We will explore the learned dynamics of the subpopulations \( C\), \( T\), and \( M\) and infer interaction dynamics modeled by the time-varying parameters \(p_C\), \(d_M\), and \(r_M\). Afterwards, we will compare both cases using metrics including mean squared error (\(MSE\)), mean absolute error (\(MAE\)), and mean absolute percentage error (\(MAPE\)) to investigate how well both cases fit the experimental data \(u_{GEM-OT1}\). 
\subsection{Case1: Learning Interaction Dynamics using 86 Interpolation data points}\label{case1} 
We utilized the Modified PINN algorithm (Section~\eqref{subSEC_mPINN}) to learn the interaction dynamics among the subpopulations \( C\), \( T\), and \( M\) in the chemo- and immune-therapy model presented in Section~\eqref{chemimmu}. First, we use spline-based interpolation to extend the experimental tumor volume data \(u_{GEM-OT1}\) from \(6\) to \(86\) data points. The interpolated data points were uniformly spaced between the experimental data points collected at days \(6, 9, 13, 16, 20\), and \(23\) (Section~\eqref{chemimmu}). We preferred the cubic-spline interpolation over other interpolation methods such as the Lagrange polynomial and the Newton polynomial because cubic-splines are smother and adds less noise to the training data for the two PINNs in Figure~\eqref{expdata}({\bf a}). Although cubic-splines need not be uniformly spaced, however, we chose the case of uniformly spaced cubic-splines in order to minimize the distance between consecutive interpolation points.

\begin{figure}[h]
	\begin{minipage}[t]{0.70\textwidth}
		\begin{minipage}[t]{0.30\textwidth}
			\includegraphics[width = 2.7in]{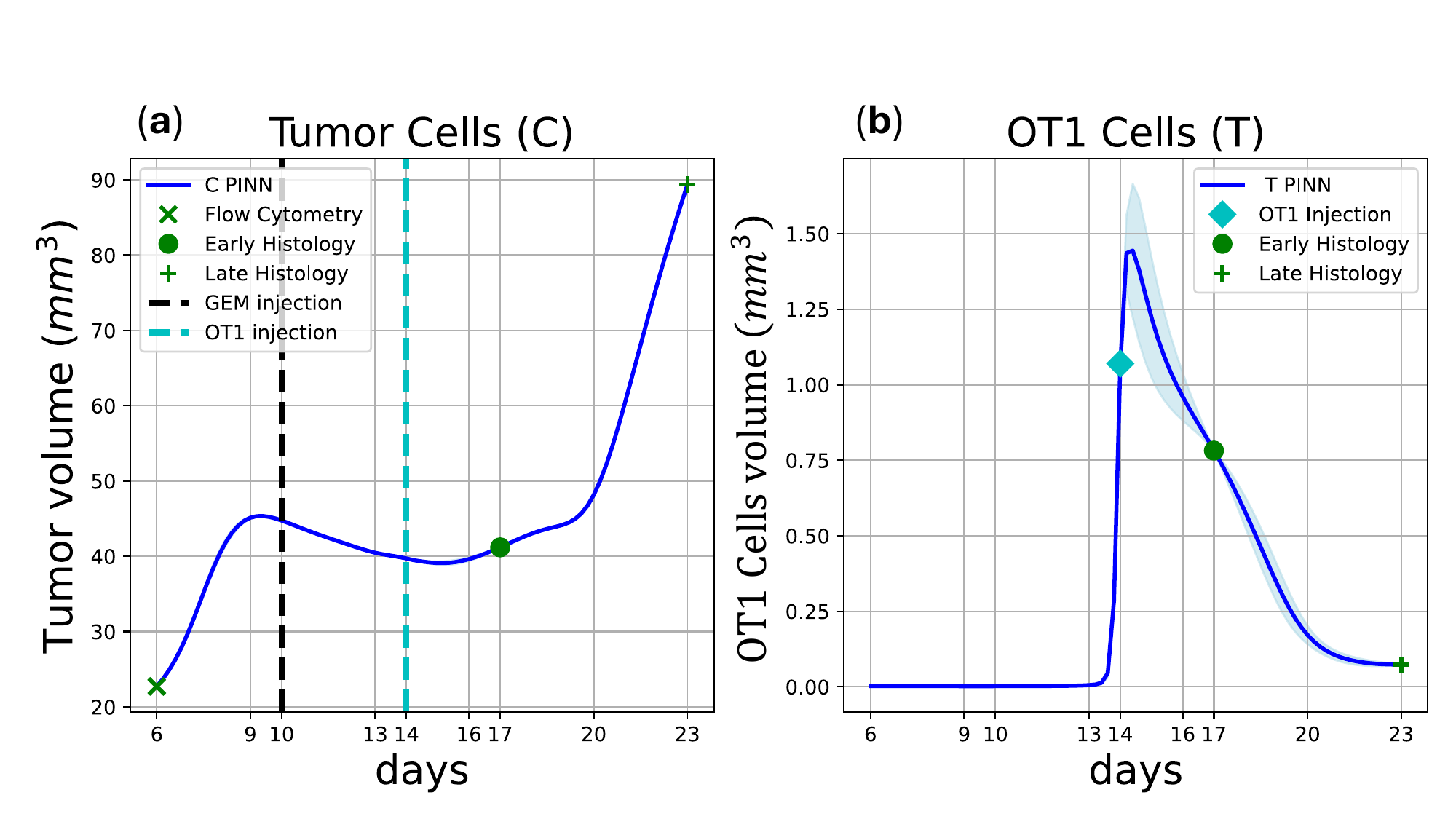}
		\end{minipage}
		\hfill
		\begin{minipage}[t]{0.30\textwidth}
			\includegraphics[width = 2.7in]{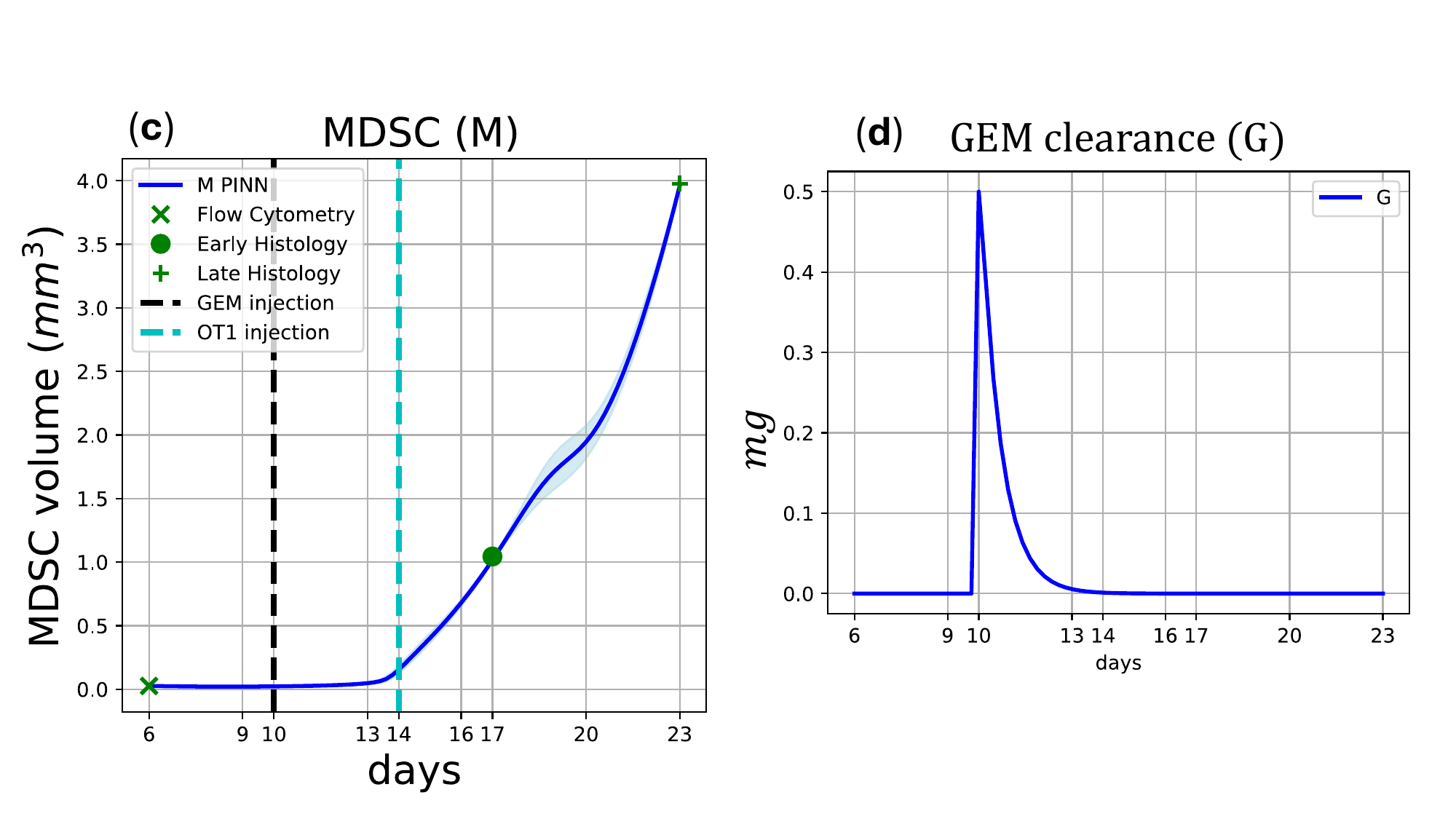}
		\end{minipage}
		
		\vspace{0.01em}
		
		\begin{minipage}[t]{0.30\textwidth}
			\includegraphics[width = 2.7in]{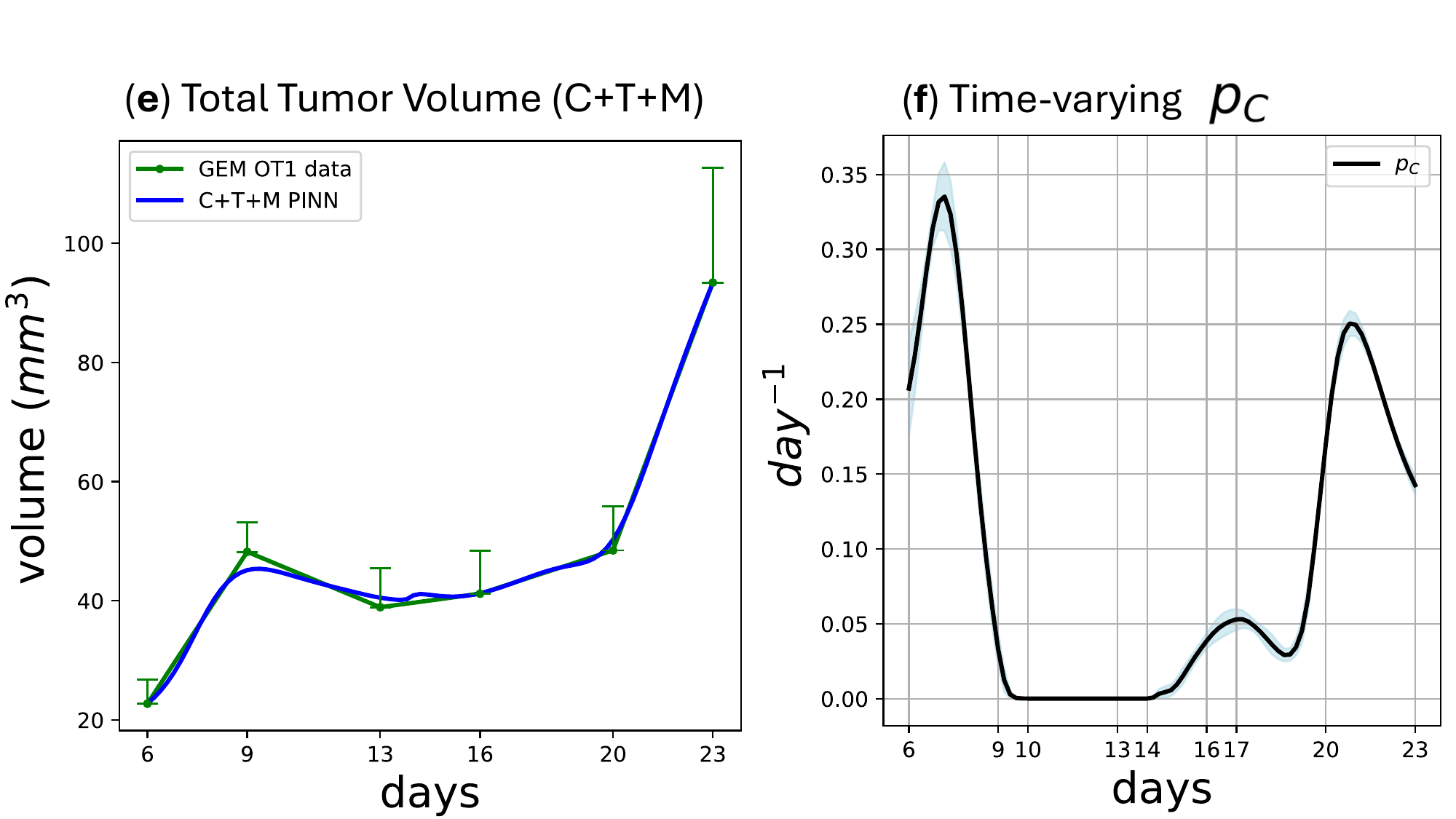}
		\end{minipage}
		\hfill
		\begin{minipage}[t]{0.30\textwidth}
			\includegraphics[width = 2.7in]{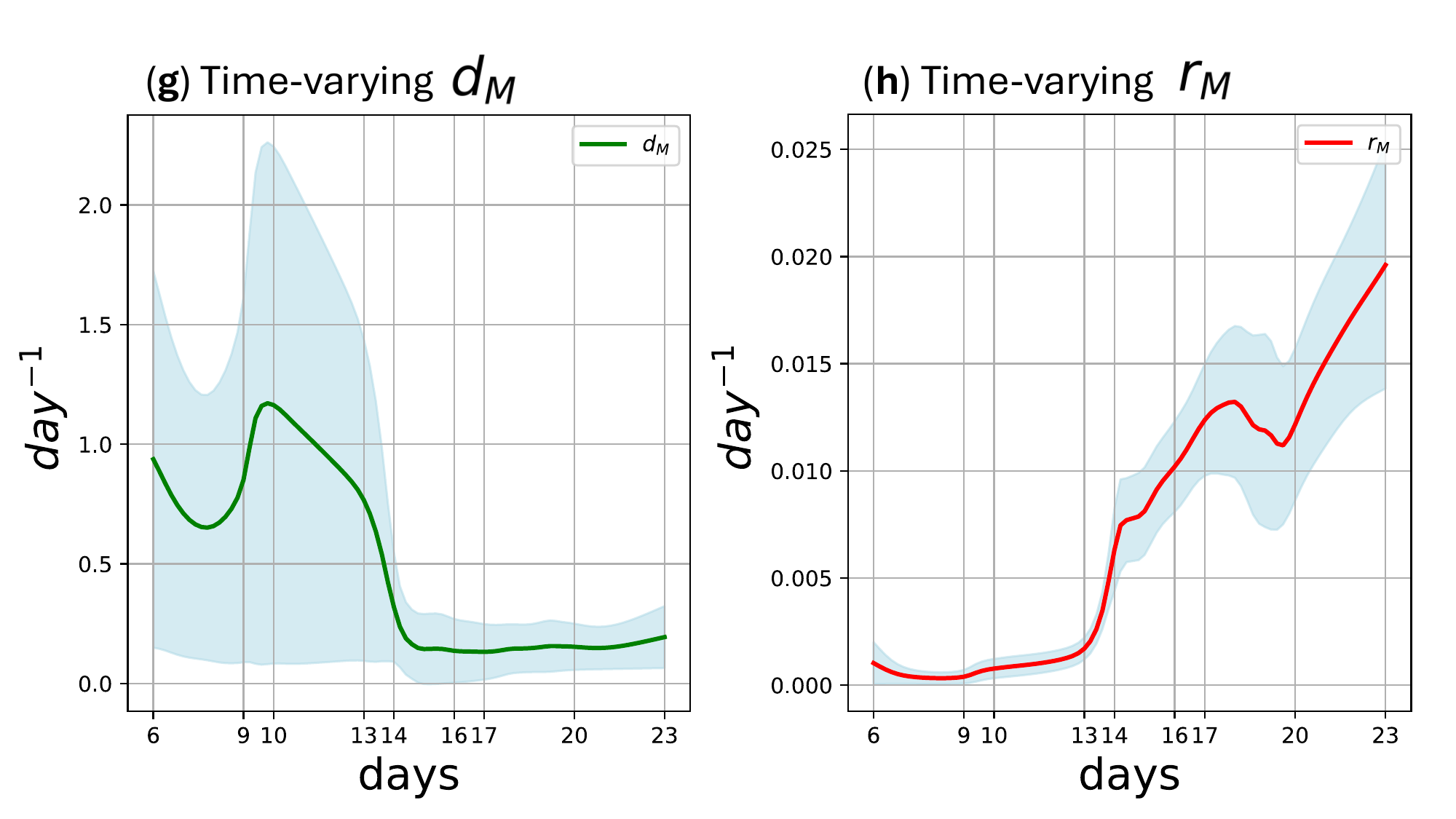}
		\end{minipage}
	\end{minipage}
\caption{Subpopulations \(C, T, M \) and time-varying \(p_C, d_M, r_M\) learned by the modified PINN using 86 interpolation data points.}
\label{subpopu1}
\end{figure}

As shown in Figures~\eqref{subpopu1}({\bf a,b,c}), the modified PINN algorithm learns a possible dynamics for the subpopulations \( C\), \( T\), and \( M\), aligning with biological constraints at the flow cytometry point (represented as green cross in Figures~\eqref{subpopu1}({\bf a,c}) and histology time points (represented as green circle dot at the early histology time point and green circle plus at the late histology time point in Figures~\eqref{subpopu1}({\bf a,b,c})). The learned clearance of the drug GEM \((G)\) is shown in Figure~\eqref{subpopu1}({\bf d}). The sum of the subpopulations \( C+T+M\) matches the experimental tumor volume data at the time points where measurements were collected (Figure~\eqref{subpopu1}({\bf e})). In Figures~\eqref{subpopu1}({\bf f,g,h}), the modified PINN learns possible interaction dynamics for the three model parameters: \(p_C\), \(d_M\), and \(r_M\) in an unsupervised fashion. After \(10\) runs of the PINN algorithm, we observe higher uncertainty in \(d_M\) than in \(p_C\) and \(r_M\) before day \(14\) where there was no OT-1 T cells \((T)\) in the model. On the contrary, \(r_M\) had the highest uncertainty after day \(14\). 
%
%
%
%
%
%
%
\subsection{Case2: Learning Interaction Dynamics using 35 Interpolation data points}\label{case2} 
%
In this case, similar to Section~\eqref{case1}, we applied the Modified PINN algorithm (Section~\eqref{subSEC_mPINN}) to the chemo- and immune-therapy model described in Section~\eqref{chemimmu}. We extend the experimental tumor volume data \(u_{GEM-OT1}\) from \(6\) to \(35\) data points, using uniformly spaced cubic-splines.

\begin{figure}[h!]
	\begin{minipage}[t]{0.70\textwidth}
		\begin{minipage}[t]{0.30\textwidth}
			\includegraphics[width = 2.7in]{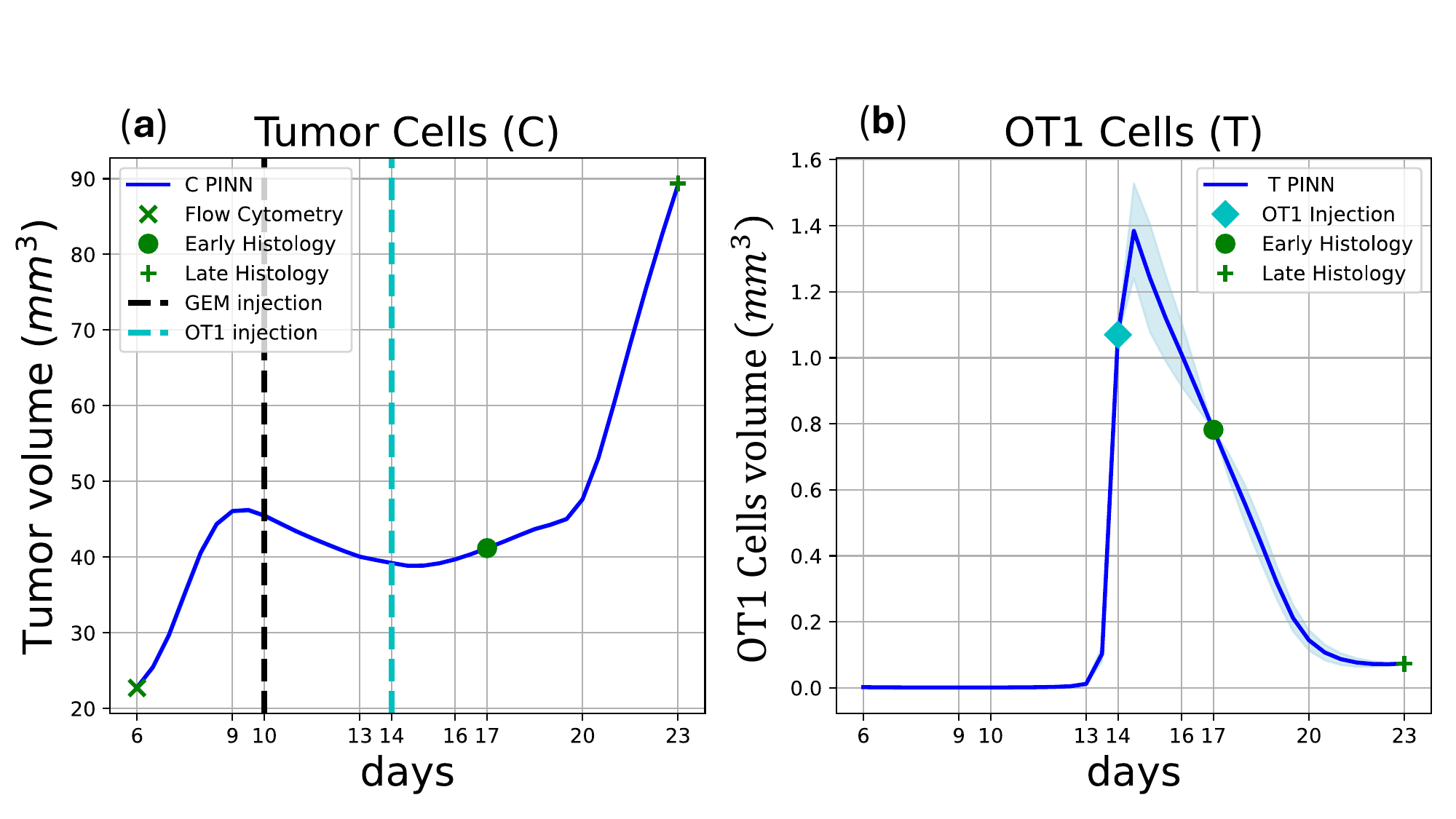}
		\end{minipage}
		\hfill
		\begin{minipage}[t]{0.30\textwidth}
			\includegraphics[width = 2.7in]{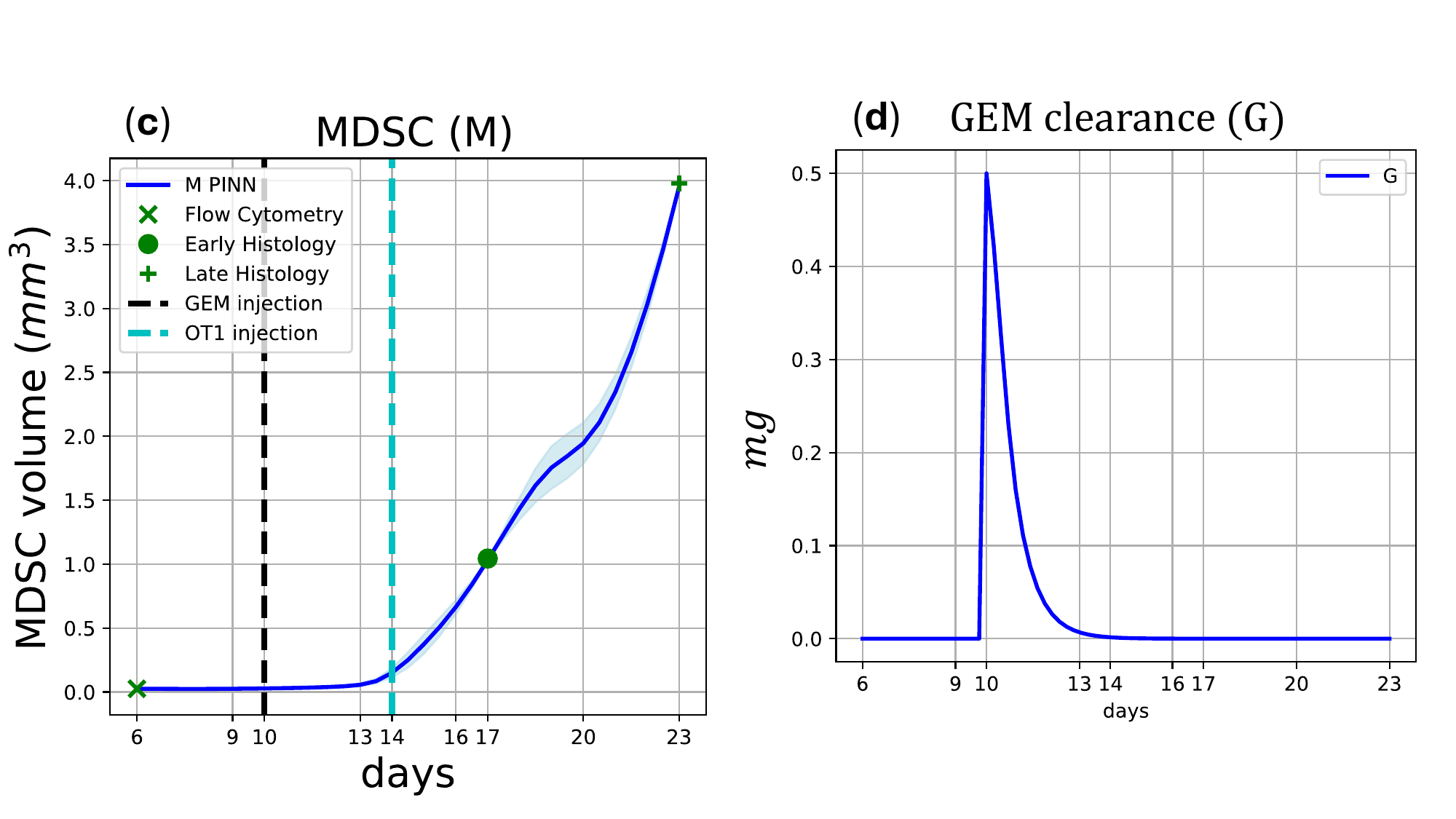}
		\end{minipage}
		
		\vspace{0.01em}
		
		\begin{minipage}[t]{0.30\textwidth}
			\includegraphics[width = 2.75in]{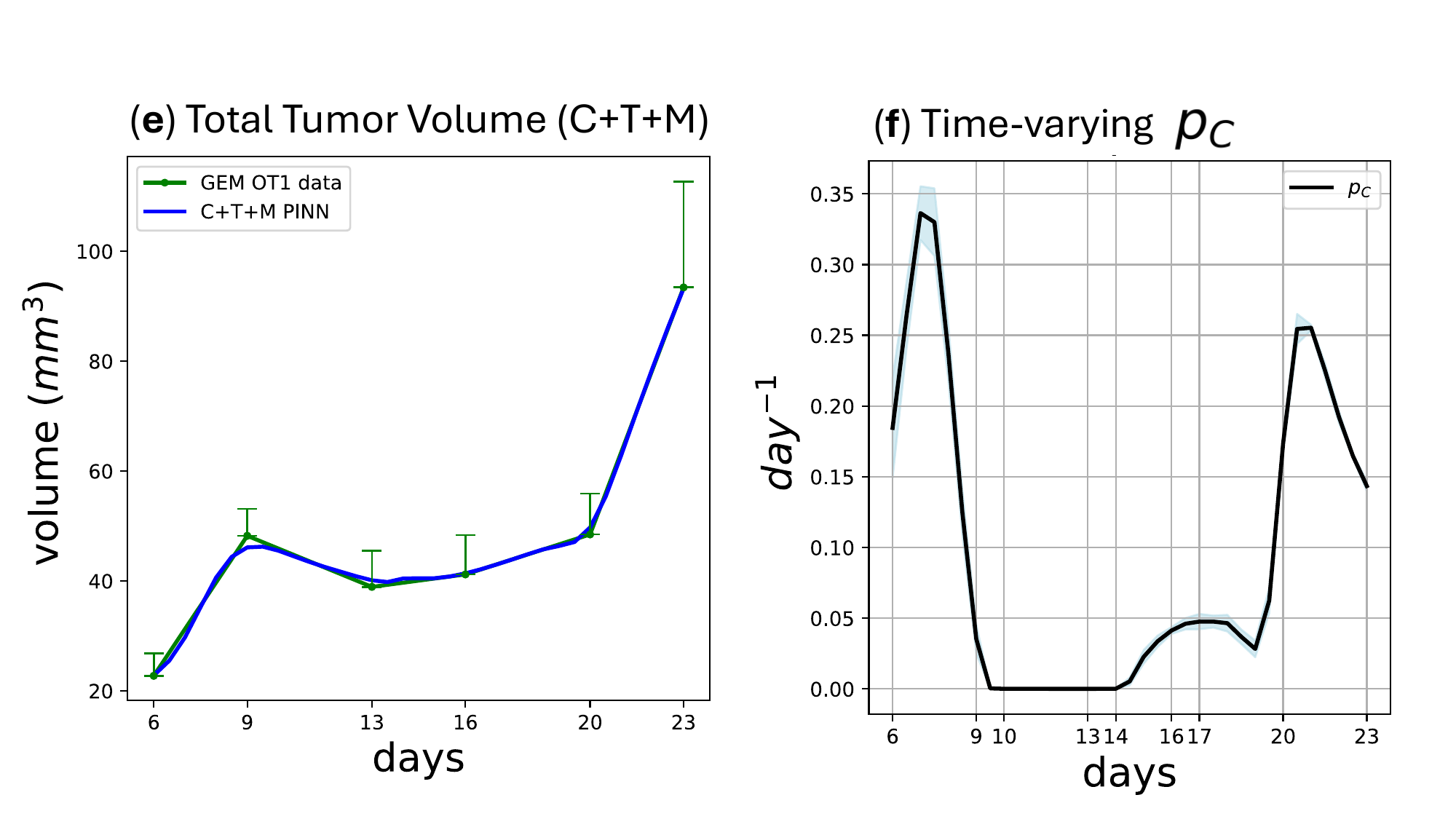}
		\end{minipage}
		\hfill
		\begin{minipage}[t]{0.30\textwidth}
			\includegraphics[width = 2.75in]{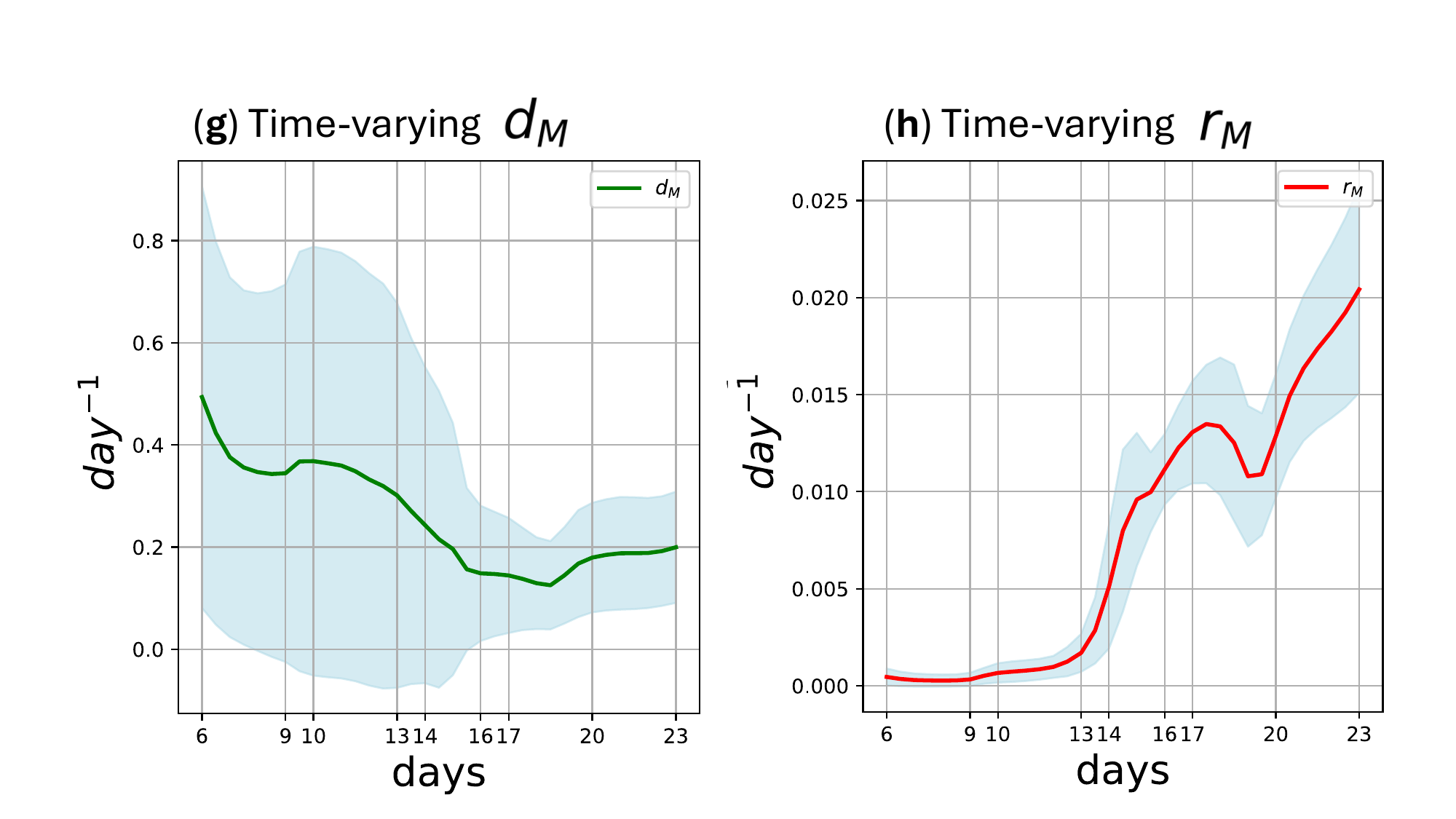}
		\end{minipage}
	\end{minipage}
\caption{Subpopulations \(C, T, M \) and time-varying \(p_C, d_M, r_M\) learned by the modified PINN using 35 interpolation data points.}
\label{subpopu2}
\end{figure}

The learned dynamics of the subpopulations \( C\), \( T\), and \( M\) are shown in Figures~\eqref{subpopu2}({\bf a,b,c}), while the learned interactions among the subpopulations due to the external chemo- and immune-therapy interventions are presented in Figures~\eqref{subpopu2}({\bf f,g,h}). The sum of the subpopulations \( C+T+M\) matches the experimental tumor volume data at the time points where measurements were collected (Figure~\eqref{subpopu2}({\bf e})) and the clearance of the drug GEM \((G)\) is shown in Figure~\eqref{subpopu2}({\bf d}). The time-varying parameter \(d_M\) has the highest uncertainty (Figure~\eqref{subpopu2}({\bf g})) when compared to the time-varying parameters \(p_C\) and \(r_M\) (Figures~\eqref{subpopu2}({\bf f,h}))  before day \(14\). However, when we compare (Figure~\eqref{subpopu1}({\bf g})) with (Figure~\eqref{subpopu2}({\bf g})), we see that the uncertainty bound is higher in (Figure~\eqref{subpopu1}({\bf g})). 
%
%
%
%
%
%
\subsection{Comparison of Loss Convergence for the two cases}
\begin{figure}[h!]
    \subfloat[86 Interpolation points]{\includegraphics[width = 3in]{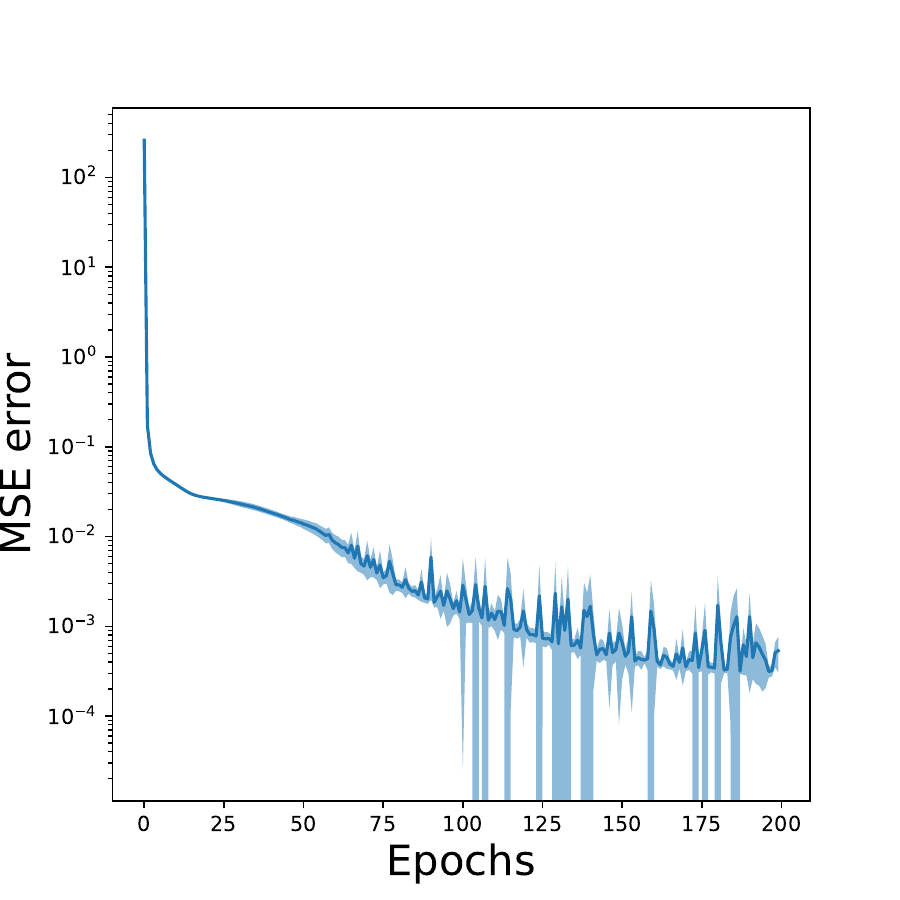}} 
    \hfill
    \subfloat[ 35 Interpolation points]{\includegraphics[width = 3in]{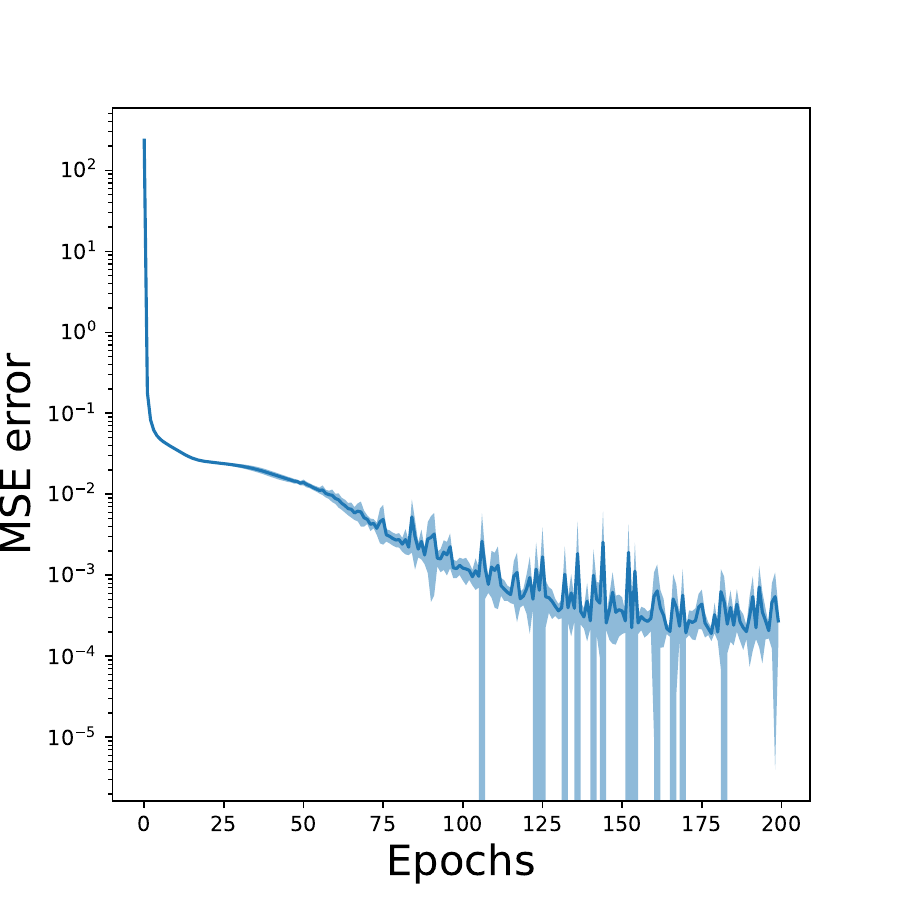}}
\caption{Convergence curve of the modified PINN for the two cases}
\label{conv}
\end{figure}

We ran the modified PINN algorithm (Section~\eqref{subSEC_mPINN}) \(10\) times for each of the cases presented in Sections~\eqref{case1} and \eqref{case2}. In all runs, we used a learning rate of \SI{e-3} in the Adams optimizer. The convergence of the training loss is shown in Figure~\eqref{conv}(a) for Sections~\eqref{case1} and in Figure~\eqref{conv}(b) for Sections~\eqref{case2}. In both cases, we observed that the loss occasionally reaches \SI{e-4} or \SI{e-5}, but it generally remains around \SI{e-3} in majority of the epochs.
\subsection{Comparison of Metrics for the two cases}
We compared the \(C + T + M\) values learned from the modified PINN with the experimental data points at the six observed time points using metrics such as mean squared error (\(MSE\)), mean absolute error (\(MAE\)), and mean absolute percentage error (\(MAPE\)), see Table~\eqref{metrictable}. We observe that Case 2 (Section~\eqref{case2}) performed better across all three metrics. We also note that the uncertainty bound is higher in case 1 (Section~\eqref{case1}) than in case 2 (Section~\eqref{case2}). This suggests that a formal investigation is needed to determine the optimal number of interpolation points required to extend the experimental data.
\begin{table}[h]
  \caption{Error metrics between experimental data and PINN predictions}
  \label{metrictable}
  \centering
  \begin{tabular}{llll}
    \toprule
    Cases     & MSE     & MAE  & MAPE \\
    \midrule
    86 Interp, time-varying & 2.5404  & 1.0987 & 0.0241     \\
    35 Interp, time-varying     & \bf{1.2629} & \bf{0.8388} & \bf{0.0184}     \\
    \bottomrule
  \end{tabular}
\end{table}
%
\section{Limitations}
Although in this work, we present an application of our algorithm to experimental tumor volume data, demonstrate its robustness in learning from limited data, and we show how to extend PINNs to incorporate biological data as a form of regularization, several areas remain for future exploration. One such direction is the use of adaptive loss weighting to address imbalances in the magnitude of different loss components during training \citep{gao2025pinn, mcclenny2022pinn}. Another unresolved question is determining how many interpolation points are necessary to effectively extend sparse experimental datasets.
\section{Conclusions}
The modified PINN approach we present in this work captures the changing dynamics in a limited-data scenario involving three different types of interacting cells and Gemcitabine. Additionally, we demonstrate how to use the PINN framework to incorporate biological knowledge (histology and flow cytometry data) as a form of regularization. This enables us to reduce the space of admissible solutions. We showed formally that the spline-based interpolation preserves convergence of the PINN solution to the experimental data. The PINN framework in this paper is applicable to many other differential equation models in biology, particularly in systems where interacting organisms alter their behavior in response to internal signals or external interventions.
\begin{ack}
This work was supported by the Department of Defense grant W81XWH-22-1-0340 and the US National Institutes of Health, National Cancer Institute grant R01-CA259387. This work was supported in part by the Shared Resources at the H. Lee Moffitt Cancer Center \& Research Institute an NCI designated Comprehensive Cancer Center under the grant P30-CA076292 from the National Institutes of Health. 
%
%
\end{ack}
%
%
\bibliographystyle{plainnat}
\bibliography{references}
\newpage
\appendix
\section{Technical Appendices and Supplementary Material}
%
\subsection{Mathematical Notations}
We introduce the symbols and mathematical notations that are used in this paper.
\begin{table}[h!]
\centering
\renewcommand{\arraystretch}{1.3}
\begin{tabular}{|c|p{10cm}|}
\hline
\textbf{Symbol} & \textbf{Description} \\
\hline
$u \in {\rm I\!R}^m$ & State variables with $m$ variables. \\
\hline
$\Lambda \in {\rm I\!R}^p$ & Set of $p$ time-varying parameters. \\
\hline
$f$ & Known nonlinear function. \\
\hline
$u_{NN}$ & Neural network surrogate of $u$. \\
\hline
$\Lambda_{NN}$ & Neural network surrogate of $\Lambda$. \\
\hline
$M^{\text{data}}$ & Number of experimental data points. \\
\hline
$\{(t_i, u_i) \}_{i=0}^{M^{\text{data}}-1} $ & Experimental data points. \\
\hline
$\hat{u} $ & A spline interpolation of the data points $\{(t_i, u_i) \}_{i=0}^{M^{\text{data}}-1}$. \\
\hline
$\hat{u}_{NN}$ & Neural network surrogate of $\hat{u} $.  \\
\hline
$\sigma_k (\cdot) $ & Activation function. \\
\hline
$NN(\cdot;\cdot)$ & A feedforward neural network. \\
\hline
$\theta$ & Set of neural network parameters. \\
\hline
$\mathcal{L}$ & Loss function used for model training. \\
\hline
\end{tabular}
\caption{Mathematical notations used in this paper.}
\label{tab:notation}
\end{table}
\setcounter{theorem}{0}
\subsection{Proof of Theorem 1}
%
%
%
\begin{theorem}[Interpolation preserves convergence of ODE solution in Eq.~\eqref{orig_eqn2}]
\label{thm:interp-ode}
Suppose that:
\begin{enumerate}
  \item[(i)] \(u,u_{NN}\in C^{k+1}([t_0,t_F];\mathbb{R}^m)\) are functions satisfying Eq.~\eqref{orig_eqn} and Eq.~\eqref{orig_eqn2}, respectively.
  \item[(ii)] The PINN surrogate parameters satisfy \(\Lambda_{NN} \to \Lambda\) uniformly on \([t_0,t_F]\), and the PINN surrogate solution satisfies
  \[
  u_{NN}\to u \quad \text{in } C^{k}([t_0,t_F];\mathbb{R}^m).
  \]
  \item[(iii)] \(I_k\) is the spline interpolation operator of degree \(k\) on a quasi-uniform mesh on \([t_0,t_F]\) with spacing \(h>0\).
  \item[(iv)] The constant $C_{\text{interp}}$ denotes the spline interpolation error constant such that
  \[
  \|I_k v - v\|_{L^\infty(t_0,t_F)} \le C_{\text{interp}} h^{k+1} \|v^{(k+1)}\|_{L^\infty(t_0,t_F)}
  \]
  for any $v \in C^{k+1}([t_0,t_F];\mathbb{R}^m)$.
\end{enumerate}
Then
\[
\|I_k u_{NN} - u\|_{L^\infty(t_0,t_F)} 
\le \|I_k u_{NN} - I_k u\|_{L^\infty(t_0,t_F)} + \|I_k u - u\|_{L^\infty(t_0,t_F)}.
\]
Moreover, as $h \to 0$:
\[
\|I_k u_{NN} - u\|_{L^\infty(t_0,t_F)} \to 0.
\]
\end{theorem}
\begin{proof}

\noindent\textbf{Part 1: Triangle inequality.}
\[
\|I_k u_{NN} - u\|_{L^\infty} = \|I_k u_{NN} - I_k u + I_k u - u\|_{L^\infty} 
\le \|I_k u_{NN} - I_k u\|_{L^\infty} + \|I_k u - u\|_{L^\infty}.
\]

\noindent\textbf{Part 2: Convergence as $h \to 0$.}
\[
\|I_k u_{NN} - u\|_{L^\infty} \le \|I_k u_{NN} - I_k u\|_{L^\infty} + \|I_k u - u\|_{L^\infty}.
\]

\noindent\textbf{Bound on the first term:}
By assumption (iv), we have
\[
\|I_k u_{NN} - I_k u\|_{L^\infty} \le C_{\text{interp}} h^{k+1} \|(u_{NN} - u)^{(k+1)}\|_{L^\infty} \le C_{\text{interp}} h^{k+1} \left(\|u_{NN}^{(k+1)}\|_{L^\infty} + \|u^{(k+1)}\|_{L^\infty}\right).
\]

\noindent\textbf{Bound on the second term:}
By assumption (iv), the standard spline interpolation error estimate gives
\[
\|I_k u - u\|_{L^\infty} \le C_{\text{interp}} h^{k+1} \|u^{(k+1)}\|_{L^\infty}.
\]

\noindent\textbf{Combining the bounds:}
From the triangle inequality and the bounds above,
\[
\|I_k u_{NN} - u\|_{L^\infty} \le \|I_k u_{NN} - I_k u\|_{L^\infty} + \|I_k u - u\|_{L^\infty} 
\le C_{\text{interp}} h^{k+1} \left(\|u_{NN}^{(k+1)}\|_{L^\infty} + 2\|u^{(k+1)}\|_{L^\infty}\right).
\]

\noindent\textbf{Convergence:}
By assumption (ii), $u_{NN} \to u$ in $C^k([t_0,t_F];\mathbb{R}^m)$, which implies $\|u_{NN} - u\|_{L^\infty} \to 0$. 
By assumption (i), both $u$ and $u_{NN}$ are in $C^{k+1}([t_0,t_F];\mathbb{R}^m)$, so $\|u^{(k+1)}\|_{L^\infty}$ and $\|u_{NN}^{(k+1)}\|_{L^\infty}$ are bounded constants independent of $h$.
Therefore, as $h \to 0$:
\[
\lim_{h \to 0} \|I_k u_{NN} - u\|_{L^\infty} = 0.
\]

\end{proof}

\end{document}